\documentclass[twocolumn,conference]{IEEEtran}
\usepackage[T1]{fontenc}
\usepackage[letterpaper]{geometry}
\usepackage{textcomp}
\usepackage{graphicx}
\usepackage[numbers]{natbib}
\usepackage[unicode=true,
 bookmarks=true,bookmarksnumbered=true,bookmarksopen=true,bookmarksopenlevel=1,
 breaklinks=false,pdfborder={0 0 0},pdfborderstyle={},backref=false,colorlinks=false]
 {hyperref}
\hypersetup{pdftitle={Your Title},
 pdfauthor={Your Name},
 pdfpagelayout=OneColumn, pdfnewwindow=true, pdfstartview=XYZ, plainpages=false}

\makeatletter
\usepackage[caption=false,font=footnotesize]{subfig}
\AtBeginDocument{%
\let\ref\autoref
\renewcommand\equationautorefname{\@gobble}
}

\makeatother

\begin{document}

\title{Towards a New Paradigm of UAV Safety}

\author{}

\author{\href{mailto:jafman3@gatech.edu}{Juan-Pablo Afman}, \href{mailto:laurent.ciarletta@loria.fr}{Laurent Ciarletta},
\href{mailto:eric.feron@aerospace.gatech.edu}{Eric Feron}, \href{mailto:jkf6x7@gatech.edu}{John Franklin},
\href{mailto:thomas.gurriet@gatech.edu }{Thomas Gurriet}, \href{mailto:eric.johnson@ae.gatech.edu }{Eric N. Johnson }}
\maketitle
\begin{abstract}
With the rising popularity of UAVs in the civilian world, we are currently
witnessing and paradim shift in terms of operational safety of flying
vehicles. Safe and ubiquitous human-system interaction shall remain
the core requirement but those prescribed in general aviation are
not adapted for UAVs. Yet we believe it is possible to leverage the
specific aspects of unmanned aviation to meet acceptable safety requirements.
We start this paper with by discussing the new operational context
of civilian UAVs. We investigate the meaning of safety in light of
this new context, keeping in mind the operational and economical constraints.
Next, we explore the different approaches to ensuring system safety
from an avionics point of view. It mostly consists in lowering failure
occurrences and managing/leveraging them. Formal verification, redundancy
and fault tolerant control are discussed and weighted against the
high cost of current methods used in commercial aviation. Subsets
of operational requirements such as geofencing or mechanical systems
for termination or impact limitation can easily be implemented. These
are presented with the goal of limiting the collateral damages of
a system failure. We then present some experimental results regarding
two of the major problems with UAVs. With actual impacts, we demonstrate
how dangerous uncontrolled crashes can be. Furthermore, with the large
number of runaway drone experiences during civilian operations, the
risk is even higher as they can travel a long way before crashing.
We provide data on such a case where the software controller is working,
keeping the UAV in the air, but the operator is unable to actually
control the system. It should be terminated! Finally, after having
analyzed the context and some actual solutions, based on a minimal
set of requirement and our own experience, we are proposing a simple
mechanical based safety system. It unequivocally terminates the flight
in the most efficient way. Our Active Cutting System instantly removes
parts of the propellers leaving a minimal lifting surface. It takes
advantage of what controllability may remain but with a deterministic
ending: a definite landing. 
\end{abstract}

\section*{Introduction}

In 2013, the number of fatal accidents in the US was down to 236 for
roughly 64 million general aviation flights. Although one should be
proud of such a safety achievement, the challenges that tomorrow will
face should be addressed in a timely and appropriate manner. Among
these challenges includes the rapid growth of the civilian UAV market,
a market that has enabled anyone who can spare a few hundred dollars
to gain full access to the general aviation airspace.

In general aviation, accidents have always had a high level of fatalities
of the people on-board (steadily around 20\% for the past 30 years
\citep{Aviation craches record}). Hence, the reason for rigorous
design and certifications processes. However, this level of requirements
costs a lot in terms of development time and vehicle cost. For UAVs,
the paradigm changes completely. With unmanned vehicles, the only
potential casualties become the people on the ground. Since human
life is at stake, these systems now become safety critical systems
that should be certified, although the same level of certification
as for general aviation would make them impractically expensive. Hence,
new strategies should be devised in order to enable proper safety
while meeting budget and design costs.

In the current state, the lack of safety measures included in current
civilian UAVs is alarming. To begin, most civilian UAVs are composed
of ``hobby-grade'' components repackaged into a shiny fuselage.
Even though some efforts have been done to implement safety features
like geofencing, the state of the art UAV remains very far from reaching
the safe-to-crash paradigm.

This paper will focus on evaluating this paradigm change that shifts
attention from crash-free, as it is in the general aviation, to safe-to-crash
design for UAVs. Furthermore, a thorough study of vehicle design and
procedures that embrace this new approach is included. The paper is
organized by first discussing this new paradigm of UAV safety in the
light of the present and future operational context, while defining
some high level safety requirements that fit this new paradigm. Then,
Focus is shifted towards hardware and software-based approaches for
meeting these safety requirements. Lastly, an innovative hardware
based feature is disclosed that enables the UAV to crash safely in
the event of the common ``run-away'' drone situation.

\section{A new operational context}

The potential applications of UAVs in the civilian community are numerous.
However, safe and reliable UAVs can only be introduced if the acquisition
and operational cost is low enough to be economically viable in today's
economy. If performance is the only cost function, it is be easy to
use low cost consumer electronics and have very cheap products. However,
the balance between performance and safety on the other hand can come
at a steep price. For general aviation, the cost of software validation
has been observed to be as high as half of the entire development
cost. Hence, the question becomes: How can we reach an equivalent
level of safety, but at a significantly lower cost? To answer this
question we need to understand the context in which UAVs evolve in
while mapping out the safety requirements which will enable civilian
applications.

Because no human being is on-board these systems, only people and
infrastructures on the ground are endangered by flying UAVs. Therefore,
the safety paradigm isn't about avoiding crashes, but about avoiding
crashing into civilians on the ground. This high level requirement
then translates into rules concerning the vehicles and the rules concerning
the operations. This two points are fundamentally coupled, and requirements
on the vehicles cannot be decided independently of the mission flown.

In this section one can distinguish two general scenarios: 1) UAVs
operating over civilians and infrastructures, or 2) UAVs flying in
un-populated areas. In the current state of regulations, civilian
UAVs are prohibited from flying over civilians and controlled airspace.
Hence, leaving room for softer requirements for the vehicles in terms
of quality and safety features. One could argue that it is a trust
based scenario where one must trust a human pilot to follow the rules.
Since strict training is not required to fly a UAV, human error often
becomes the main cause for UAV disasters. Furthermore, due to the
lack of presence of pilots on the UAV, it becomes more difficult to
detect and react to failures. Therefore, it is believed that mandatory
efforts should be made to embedded decision making features into avionics
in order to ensure the enforcement of the rules at all times. 

Since the safety requirements are strictly mission specific in UAV
applications, it is important to understand the operational context
each UAV will be subjected to in order to design a vehicle with the
appropriate safety features. Having wings to glide to a safe area
could for example be mandatory if flying over a crowd, whereas a kill
switch, instead of wings, could be required when flying over un-populated
fields to prevent the UAV from drifting into a nearby schoolyard in
the event of a communication failure. In the end, a lot of safety
enhancing features can be incorporated intelligently into UAV to reduce
the risks and minimize the damage in the event of a failure.

As society continues to strive forward, the level of autonomy in UAVs
will increase proportionally. Eventually, one can imagine reaching
a state where operation becomes so complex and fast paced that we
would reach the limits of humans capacity. Hence, a sole reliance
on automated control algorithms. In order for this scenario to reach
full maturity, a level of ``automated safety'' must evolve. In that
futuristic but not fairy context, human-system interfaces will play
an increasingly important role as pilots operate UAVs at higher levels
of abstractions. Eventually, the range of the scale at which UAVs
operate will be expanded, from advertising inside a mall to long range
cargo transport. Therefore, the entire notion of airspace will require
once again a revision in order to take these different scenarios into
account where the dynamic nature of the interaction between UAVs,
people and the environment become important. In the end, it is about
this entire infrastructure that must be built around the emerging
UAV market, and making this transition to a fully automated airspace
will only be possible if safety is at the core of this revolution.

\section{Safe-to-crash uav design: An Avionics Perspective}

\subsection{Enforcing flight safety}

As discussed in the introduction, automation can ensure the safe operation
of a UAV. In practice, this means making sure it only flies in specific
areas defined by the legislation or by the operator. In this context,
the concept of geofencing becomes important. Similar to safety guards
on interstates that prevent a car from deviating into the in-coming
traffic, a geofence defines the operational ``safe'' regions of
space. Although geofencing is commonly used by defining a safe-to-fly
space using GPS coordinates, one can further define a safe sets on
the entire state space of the vehicle as well as its flight envelope.

To address this problem, different ideas have been explored. A lot
of them fall into the path-planning category. The idea is to compute
a path in a know environment that gets the UAV to a desired position.
This can be done by solving optimization problems\citep{MIP path planning}
or using randomized algorithms\citep{RRT proba path planning}, and
if we now have the computational power to do that in real time, it
is still difficult to get formal guaranties of safety\citep{MIP path planning}.
But what happens when it is a human controlling the system as it is
the case for most UAVs today? In that context, path planning algorithms
fail to provide an operational solution. This is where an second kind
of ``obstacle avoidance'' algorithm enter into play. The idea is
to operate directly inside the control loops of the systems and enforce
safety through real-time feedback. That way, the safety is assured
independently of where the input comes from (human pilot or high level
controller). Several methods can be used to realize this approach
\citep{Simplex} and we are currently working on an optimization based
solution to this problem \citep{ICUAS geofencing,Barrier}.

\subsection{Managing failures}

Detecting and managing faults on a UAV is a task that must be performed
by the autopilot. Therefore, fault detection becomes an essential
part of this type of safety-critical system. For years, several methods
have been proposed for detecting possible issues in dynamic systems
to anticipate the loss of functionality. Although most of these methods
are more suitable for off-line fault detection tests \citep{basseville,esna},
the rapid evolution of digital computers and micro controllers over
the past decade has enabled the implementation of online-fault-detection
algorithms either through physical redundancies or software based
algorithms \citep{ding,patton,isermann}.

\subsubsection{Redundancies}

\begin{figure}[tbh]
\centering{}\emph{\includegraphics[width=0.5\columnwidth]{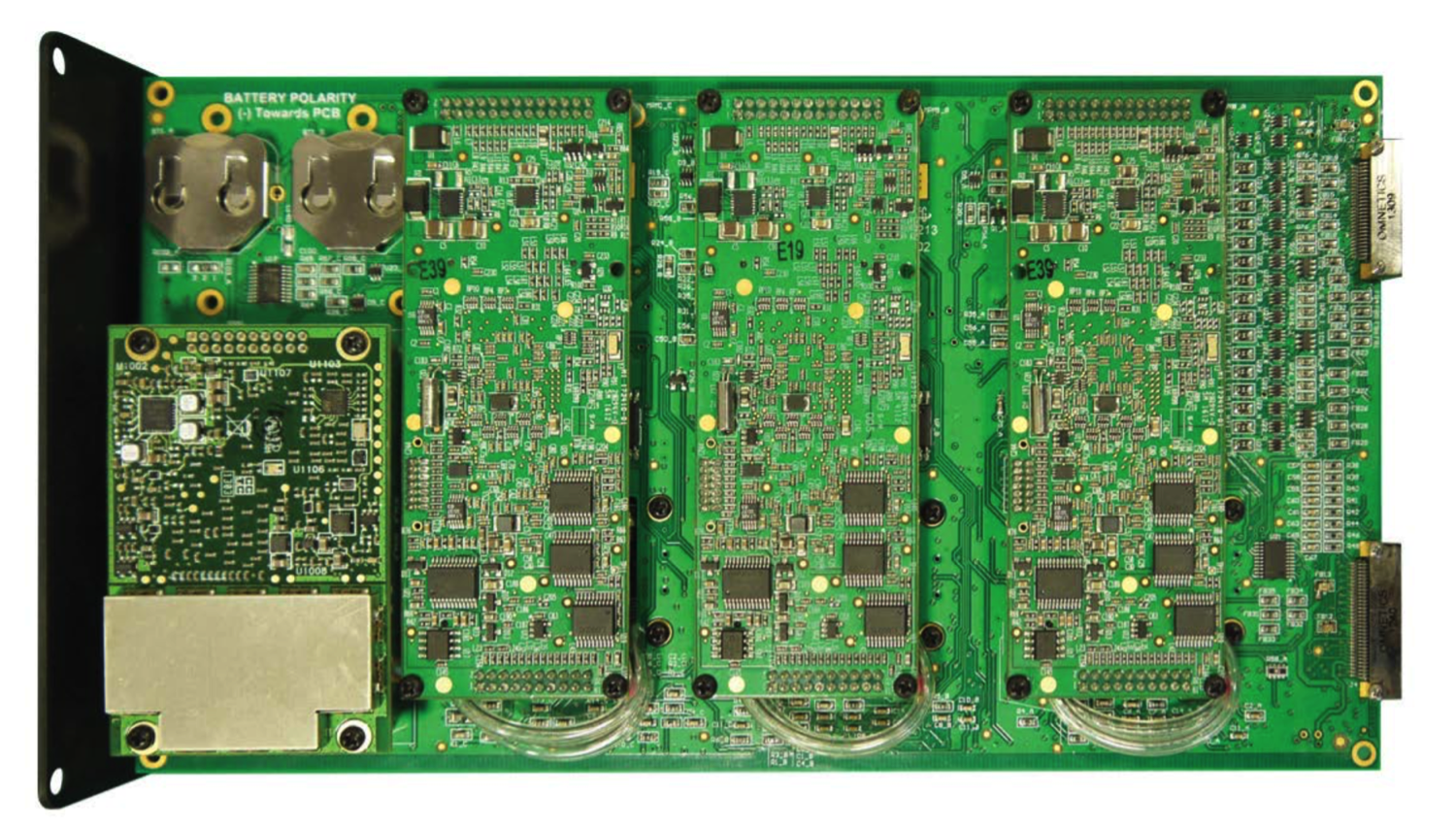}}\caption{The MP2128-3X contains three MP2128-HELI2 autopilots \citep{micropilot}
\label{fig:The--contains}}
\end{figure}

Fault management through physical redundancies allow autopilot technology
the reliability necessary to safely carry out sensitive flight missions
by incorporating multiple autopilots into one device. An $n^{th}$
redundant arrangement is comprised of n similar software and hardware
systems. If any one of the $n$ systems fails, the remaining $n-1$
continue the operation. Note that an additional ``voter'' mechanism
is also included to oversee these systems. Although redundant autopilots
are not new and well established within the aviation industry, military
aircraft such as the RAF\textquoteright s Trident fleet used a triple
redundant autoland systems in the early 1960\textquoteright s, and
ten years later, the Aérospatiale-BAC Concorde took advantage of 3X
technology in its flight control system, redundant autopilots are
a relatively new addition to UAVs. MicroPilot\textregistered , one
of the leading professional grade UAV autopilot manufacturers, set
the benchmark for triple redundancy UAV autopilots when it launched
the MP2128-3x at the end of 2010. \ref{fig:The--contains} illustrates
this device. Since the autopilots do not have an individual casing,
the overall weight and real estate added is kept to a minimum. However,
the one major drawback in this case belongs to the financial set,
since each one of these units retails for over \$6,000 USD. 

Fortunately, high-end ``hobby-grade'' flight control boards are
emerging with implemented redundancy features at significantly lower
costs. For example, the Pixhawk \citep{pixhawk} flight control board
manufactured by 3D Robotics\textregistered , features advanced processor
and sensor technology from ST Microelectronics\textregistered{} and
a NuttX real-time operating system which delivers incredible performance,
flexibility, and reliability for controlling any autonomous vehicle
at a cost of \$200 USD. It uses advanced algorithms in order to perform
\textquotedblleft sanity checks,\textquotedblright{} which compare
data from different sensors and ignore figures that seem inaccurate.
It contains a dual IMU system where an Invensense MPU 6000 supplements
ST Micro LSM303D accelerometer to provide redundancy and improve noise
immunity of the power supplies. Furthermore, it is triple-redundant
on the power supply if three power sources are supplied and will automatically
switch in the event of failure.

\subsubsection{Software Based Fault Management}

\begin{figure}[tbh]
\begin{centering}
\includegraphics[width=0.75\columnwidth]{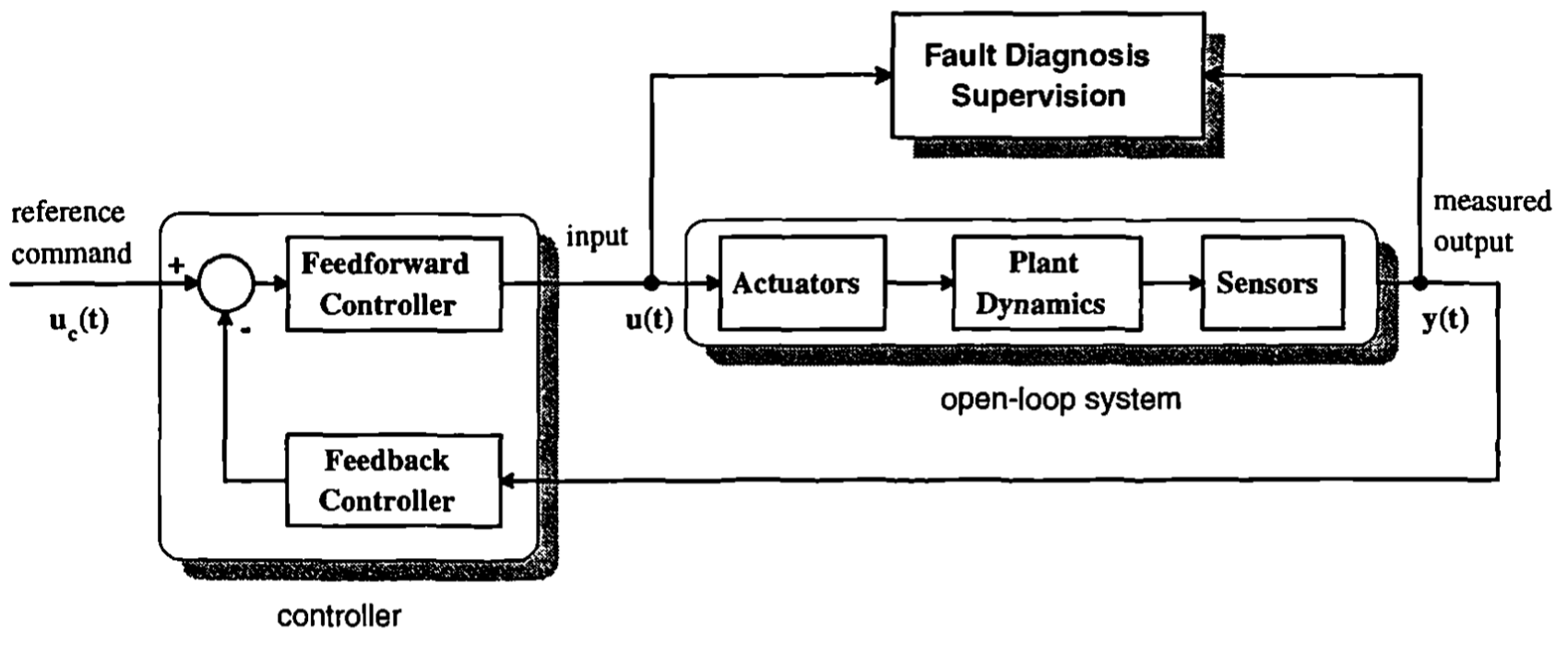}
\par\end{centering}
\caption{Open loop structure of a full-order unknown input observer \citep{patton}\label{fig:Open-loop-structure}}
\end{figure}

As we have seen in the previous section, professional grade autopilots
are often outside of a typical budget. Furthermore, depending on their
size, they could have a significant effect on the performance of the
vehicle and could decrease the payload the UAV can carry. One of the
benefits of online software-based fault management is that it can
be carried out at very little added computational cost without affecting
the controlling algorithms. Software based fault detection and management
provides analytical redundancies for monitoring the operation of the
system, analyzing input-output data and comparing the result with
the nominal behavior of the system. Some of these methods often go
as far as isolating faults or even estimating their degree of severity
\citep{ding,patton,isermann}. Among various model-based fault detection
methods, observer-based methods have been studied more than the others.
They have been implemented in industry in the past for the monitoring
safety-critical systems \citep{marzat,fekih}. Furthermore, a survey
on those methods is included in \citep{hwang}. One particularly simple
software based fault management algorithm is the Unknown Input Observer
(UIO). UIOs provide two important functional properties related to
UAV safety: detecting and isolating faults. Arguably, the greatest
benefit from this type of software-based fault detection and isolation
algorithms is that it operates in open loop mode. Hence, it does not
intrude nor affect the vehicle's control law since it only uses the
system's inputs and outputs to perform its duties. \ref{fig:Open-loop-structure}
illustrates that the system model required in model-based fault detection
and isolation is the open-loop system model although we consider that
the system is in the control loop. This is because the input and output
information required in model-based fault detection and isolation
is related to the open-loop system. Hence, it is not necessary to
consider the controller in the design of a fault diagnosis scheme. 

\subsection{Software Reliability and Validation}

The operational requirements for UAVs require the adoption of advanced
and intelligent control algorithms. In order to meet safety and operational
demands, significant progress has been made towards the development
of control and health management algorithms with features like dynamic
programming, online learning and adaptation, self-tuning and reconfiguration.
However, the use of these advance control algorithms is limited due
to the fact that flight-certification of these systems requires that
they undergo thorough Verification and Validation V\&V to achieve
high confidence in their safety. Unfortunately, the certification
of these algorithms has proven to be sometimes impossible with the
current state of the art V\&V practices, not to mention the immense
costs associated with such task. In the current state of V\&V, certification
is highly dependent on exhaustive testing from Monte-Carlo simulations
to ensure proper functionality across the flight envelope. However,
with the increase complexity of these algorithms, these methodologies
for V\&V become prohibitively costly and in some cases unfeasible
at achieving a proper level of safety confidence. There are efforts
underway to improve the practical applicability of design-time V\&V
approaches based on formal methods like theorem proving, model checking,
to provide mathematical proof of the safe execution of highly complex
advanced systems. The V\&V of all these algorithms and their implementation
could be similar to today's industry standards, but the associated
costs should be measured against the overall UAV market value for
the industry to remain competitive. Minimizing the cost of system
certification via extensive process automation and component-based
system safety evaluation are definitely two of the main challenges
in this area. For example, changing a propeller should not void the
safety certificate of the UAV.

\subsubsection{\emph{Credible Autocoding}}

Simulations and real world practical tests on system are necessary.
However, we can never test all possibilities for input signals and
fault scenarios. In addition, the software might work slightly differently
from what we expect in theory and based on original design specifications,
due to the computational errors. To detect hidden bugs and errors
we need to perform many tests. Still, the direct link between software
operation and the original mathematical proofs is missing. This issue
is more crucial for fault detection in safety-critical systems like
UAVs. In the light of evolving software certification requirements,
it becomes important to formally specify the correctness of these
algorithms.

One way of developing reliable software for safety-critical systems
is by using formal methods, which are mathematically based languages,
techniques, and tools for specifying and verifying such systems \citep{clarke1}.
Static software analysis methods include \textquotedblleft model checking\textquotedblright ,
\textquotedblleft theorem proving\textquotedblright{} and \textquotedblleft abstract
interpretation\textquotedblright , where the latter can be sometimes
interpreted as one instance of deductive software verification \citep{peled2}.
Credible autocoding becomes a complement to these methods as it focuses
on software design and the ability to insert software semantics at
design and coding time as opposed to a-posteriori semantics extraction.
The semantics included in the code look very much like those found
in deductive sequential software verification (See \citep{peled2},
Ch. 7).

\begin{figure}[tbh]
\begin{centering}
\includegraphics[width=0.6\columnwidth]{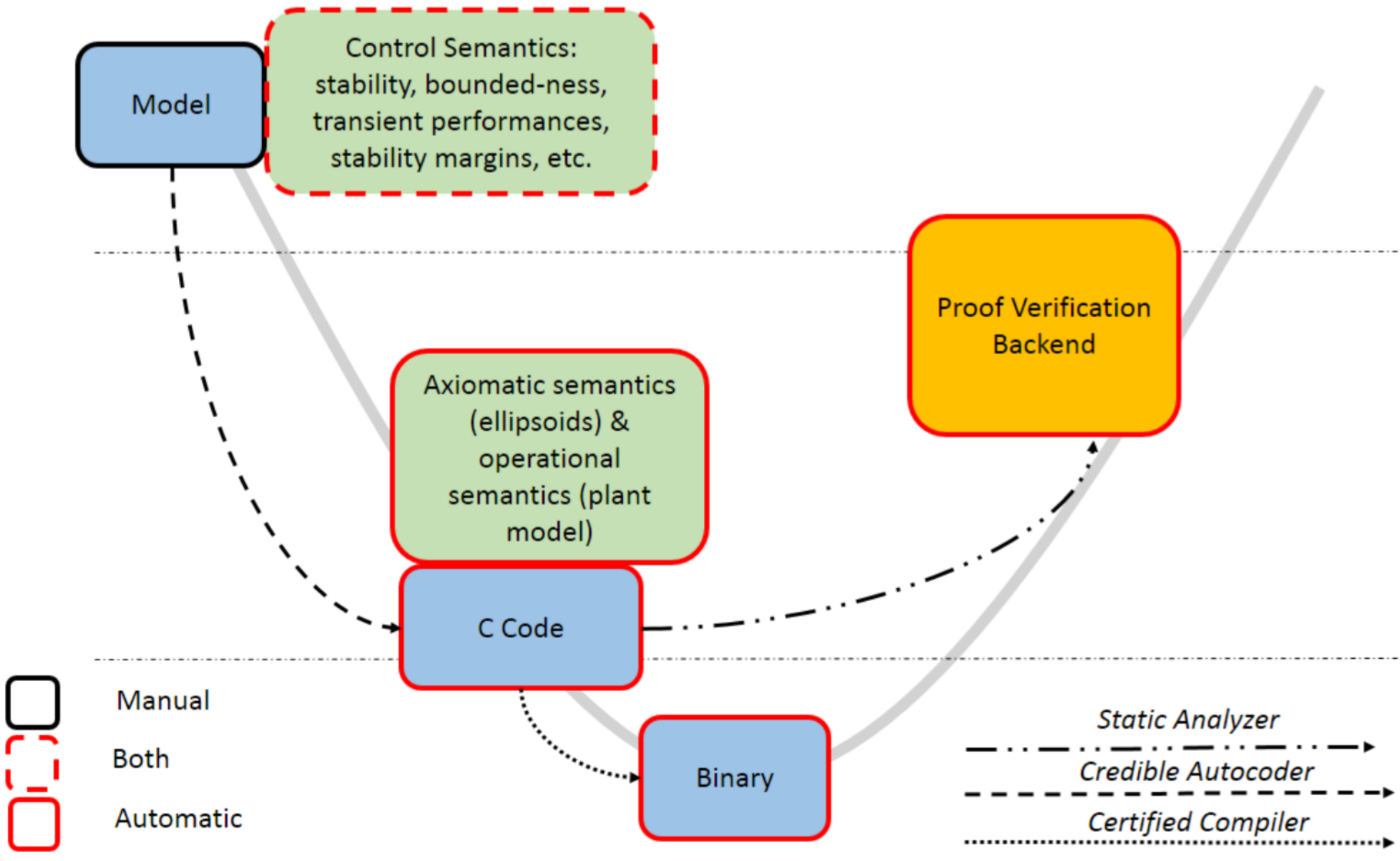}
\par\end{centering}
\caption{Schematic of Credible Autocoding Chain \citep{feron5} \label{fig:Schematic-of-Credible}}
\end{figure}

A credible autocoding tool-chain prototype based on theorem proving
approach had been built to automatically generate annotated control
software and then verifying them \citep{wang3,feron5}. In \citep{wang3,feron5}
the aim is to verify the stability of the controller and closed-loop
system controlled by software. For that purpose, annotations are generated
along with the control software so that the control software can be
automatically analyzed by theorem proving tools. Invariant sets are
chosen for system states based on Lyapunov theory. A theorem prover
can check the validity of those invariant sets automatically provided
that it is equipped with the necessary mathematical theories and strategies.
If the sets are proved to be invariant by such theorem prover, the
software is verified. \ref{fig:Schematic-of-Credible} illustrates
the credible autocoding chain.

However, it is uncertain whether these methods will address all of
the difficulties associated with achieving the necessary confidence
in the use of these algorithms in safety-critical systems. In particular,
algorithms that are adaptive, reconfigurable or non-deterministic
in nature present the greatest challenge to design-time verification
approaches. While there are also current and ongoing efforts to develop
analytical proofs and stability/convergence guarantees for some of
these algorithms, often the assurance results are deemed relatively
weak and insufficient in meeting the stringent criteria for safety-critical
purposes.

\subsubsection{\emph{Run Time Assurance for Complex Autonomy}}

\begin{figure}[tbh]
\centering{}\includegraphics[width=0.75\columnwidth]{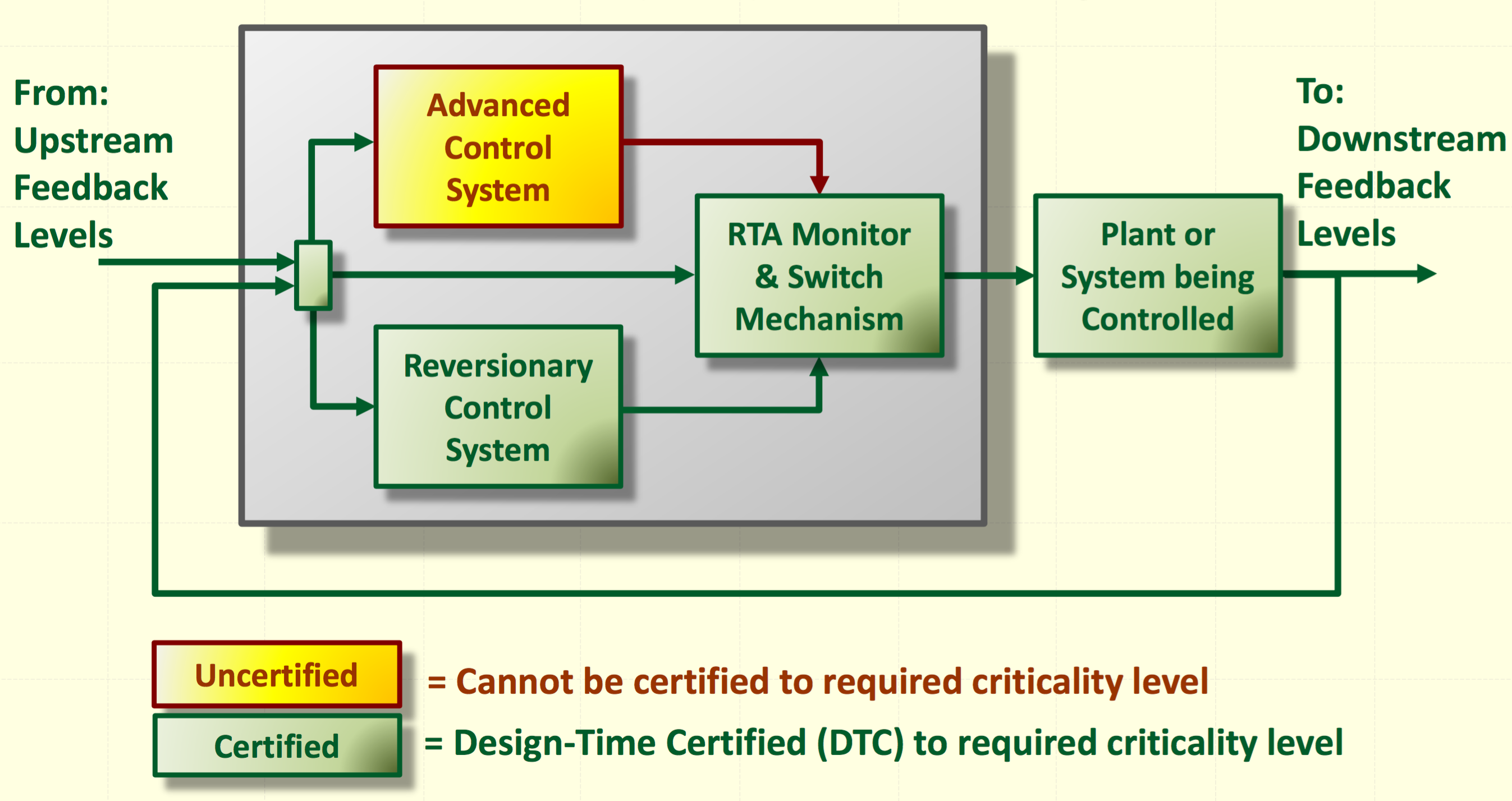}\caption{Run-Time Assurance Framework\label{fig:Run-Time-Assurance-Framework}
\citep{Schierman}}
\end{figure}

Since there exists a growing realization that no single V\&V approach
will be sufficient to address all of the challenges associated with
providing safety guarantees for these advanced or complex systems,
it is speculated among the community that run-time monitoring or run-time
verification methods can play a complementary and enabling role \citep{Schierman}.
The basic premise makes use of \textquotedblleft monitors\textquotedblright{}
to observe the execution of an uncertified algorithm in question to
insure that resulting system behavior remains constrained within acceptable
bounds of stability. A schematic of the run-time assurance methodology
is illustrated through \ref{fig:Run-Time-Assurance-Framework}.

In a run-time assurance framework, the primary advanced algorithm
is responsible for achieving all performance objectives. These algorithms
can be non-deterministic, adaptive, and enabled at all times under
nominal conditions even though it is difficult to fully certify at
design time. However, the backup system (Fail-Safe) hides in the background
and is composed of a simplified control system with emphasis on safety
rather than performance. It does not posses advanced features that
are difficult and expensive to certify. Hence, this control law is
certified at design time using state of the art traditional methods.
The RTA monitor and transition control continually monitors the overall
state of the system, including critical parameters such as safety
and operational limits, as it compares against validated representation
of safe operating envelope. If a violation occurs, the transition
controller disables the advance system and transfers control back
to the backup system.

\subsection{Security}

One important issue with UAVs is that, not only should the systems
be safe and reliable and the operators trained and respectful of the
laws, regulations and common sense, but should also be protected against
virtual ``attacks''. It should be noted that these are also counter-measures
that are being explored by the different security authorities around
the globe as a way to control the threat that UAVs can represent.

Since they work on unlicensed bands and standard technologies it is
quite easy to physically disrupt the communication channels used for
controlling the drone or for its telemetry. This can lead to some
sort of simple denial of services which will eventually cause the
UAV to enter a safe mode (Return to Home, Hover, Land, Safe Crash...).
Now someone can purposely cut the communications of a UAV. Depending
on its fail-safe program, he could therefore be taking down the system,
which would hover and eventually land, to grab its payload, or simply
steal the expensive UAV. In our crowded ``airwave'' space, it would
not require much to achieve such results. The RC world mitigates the
``noise'' issue with things like frequency/channel hopping or diversity,
which copes with the natural growth of the number of objects flying
at the same time and emitting simultaneously. The notion of coexistence
as it was studied between WiFi and Bluetooth is a challenge if you
extend the number of competing and incompatible technologies and users. 

Of course, the use of Return to Home mode needs a working GPS. With
the advances in cheap electronic (COTS), GPS scrambler units (even
though they are illegal in some countries, we are not here considering
people that abide the law) or Software Defined Radio, it is feasible
to make the RTH difficult or impossible: the issue is eventually a
crash or un-managed landing. The same goes with the use of cellular
(3G/4G etc.) modems which could also be scrambled even thought their
operating frequencies are licensed.

But with the use of general RC communication technologies or more
domestic wireless ones such as Bluetooth or WiFi, or analog video,
one could conduct more advances cyber attacks such as trying to gain
access to the telemetry or the video feeds, getting data and even
taking the control of part of the remote systems. 

Examples of hijacking drones have been studied and published\citep{HackingProDrone}.
Even without having access to the embedded software, remote activation
and destabilization could be the result of well thought fuzzing for
example. The implementations of all the protocols used in these systems
should be tested against such attacks.

It will eventually be necessary to protect the code running in the
Ground Stations and in the UAVs. Since most of the Ground Stations
are installed on more general systems (PCs, SmartPhones), they could
be used as Trojans and leave backdoor to access the data and controls
of the drone. When updating the firmware by downloading new ones from
the internet, the same type of attacks could be performed. In the
same principle as the one used to ensure that the controller of the
UAV behave correctly, supervision could be used to monitor the software
behavior of the controller, i.e for example the standard succession
of system calls \citep{ICUAS_OS_Security}. 

Recently, proof of concepts of very advanced attacks such as GPS spoofing
have made the headlines \citep{GPS_Spoofing,GPS_Spoofing2}. Using
this, a UAV path could be diverted to any other location the attacker
would want to. Regarding security and safety in general, the last
important point is to develop forensic for the recovered drones and
ground stations so that information such as proofs of wrongdoing and
identification of the wrongdoers can be collected.

\section{Safe-to-crash uav design: A Mechanical Design Perspective}

Up to this point, with the exception of redundant control boards,
software-based approaches for making UAVs safer has been the focus
of this article. As the avionics become increasingly involved, the
cost of developing and maintaining overly complex software may reach
its limit and become prohibitive. Furthermore, every effort related
to V\&V can only allow us to asymptotically converge to an ideal safety
state, but the remaining tolerance has to be handled with intelligent
system design of the vehicle itself. Hence, in some cases it may be
more efficient to have simpler software combined with efficient safety
features that are implemented at the hardware level. This approach
is coherent with the current regulation mindset that tolerates crashes
as long as they are safe for people on the ground. In what follows,
several hardware solutions will be discussed that could be used for
civilian UAVs, with special emphasis on the growing multi-rotors scene
as they seem to be the most popular platform employed for commercial
applications.

\subsection{Reducing impact energy}

As discussed earlier, the main safety requirement is about humans.
Therefore, most safety specifications are based on potential injuries
to the human body. As the head is naturally exposed to vehicles falling
from the sky, blunt head trauma is one of the most likely injury that
can have devastating short and long term effects \citep{Head trauma}.
Therefore, reducing the projectile impact energy, in this case the
falling UAV, is the aim of this topic. The French regulatory agency,
for example, limits the allowable impact energy to 69 Joules\citep{Arr=0000E9t=0000E9 UAV france}.

\subsubsection{Parachutes}

The most common device actually used for UAVs is the parachute. Since
its popularization during WWI, this technology has matured and is
now a relatively reliable solution for slowing down objects. However,
in order for a parachute to have time and open, a minimum flight altitude
is required. If maximum altitudes are explicitly defined by most regulations,
it is down to the operator to fly its UAV in order to assure the proper
functioning of the device in order to limit impact energy. This is
because most regulations prevent UAVs flights over populated area,
but as we move forward, rules similar to general aviation will have
to be put in place regarding flying over populated areas. These rules
should integrate the recovery systems capabilities of current UAVs
and maybe impose what type of technology to use. Predefined take-off
and landing site may also be part of the solution to safely get to
the minimum altitude.

Another critical component to parachute operation is its passive nature
when it comes to wind. The highest it is deployed, the more uncertain
the landing spot becomes\citep{Parachutes for UAVs}. This is true
in the event of an engine jam at full throttle on a run-away UAV.
Therefore, deploying the parachute as soon as possible may not be
the best strategy. Furthermore, even if the parachute is deployed
in the proper conditions and timing, there exists a significant number
of ways in which the canopy and/or the lines can malfunction or get
stuck on a fly-away drone whose attitude is not necessarily compatible
with parachute deployment. Note that parachutes need to be inspected
and re-folded regularly generating room for error, especially if this
task is not performed by a professional. In the end, parachutes, if
use correctly, can be efficient devices. However, due to the low tolerance
for failure, imposing regulations on the technology itself seems necessary,
as is the usage of complementary technologies to reduce the impact
trauma on the surrounding civilians.

\subsubsection{Lifting control surfaces}

As we discussed previously, a major drawbacks for parachutes is the
passive, often uncontrolled, nature of the fall. Controllable para-foils
(or ram-air) can be used in exchange, but with the same other drawback
of a parachutes. Therefore, the use of failure-dedicated control surfaces
as an integrated part of the system becomes worthy of discussion.
Such a device could allow a failed system to navigate away from a
crowd and is likely to reduce the impact energy. A thorough investigation
revealed that similar concepts have been developed in recent years\citep{Free fall cam}.
Furthermore, it was noted that the real-state impact and weight associated
could be kept at minimum on a classical multi-rotor.

In complement to control surfaces, a lifting component could be added
to better slow down the fall. A similar concept made the dynamic stability
of the Virgin Galactic SpaceShipTwo system possible. To conclude,
the need for hybrid vehicles that combine wings and stationary flight
capabilities is already present, and it is simply a matter of time
before purely vertical flight UAVs such as the classical multi-rotors
become a thing of the past. 

\subsubsection{Controlled disintegration}

Parachutes and lifting surfaces mainly aim at lowering impact velocity,
but it has been discussed that another way to minimize impact energy
would be to reduce the UAVs mass through a controlled disintegration.
Strategically destroying the vehicle is definitely not appropriate
for general aviation where the safety of the people on board is the
priority, but with UAVs it becomes a interesting and viable option.
This option has been discussed in the distant but not antithetical
context of asteroid deflection \citep{Exploding asteroid} where transforming
a big mass into a cloud of smaller debris allows for a better dissipation
of energy into the atmosphere. 

If the explosion is quite dramatic for asteroid deflection (nuclear
explosion!), it can be applied in a much more controlled and safe
way for UAV. Polymer-bonded explosives PBX for example, exhibit good
strength and machining capabilities\citep{PBX CNC,PBX strenght}.
This material could therefore be used for making specific parts of
the vehicle to provide controlled destruction capabilities of the
UAV. Sequential destruction strategies could then be though of to
intelligently reduce the vehicle into smaller pieces, possibly through
a chain reaction as is done in building demolition\citep{Building destruction}
or rocket stage separation\citep{Pirotechnic in aerospace}. Despite
they intimidating nature, pyrotechnics are now a well mastered technology
that the general public is subjected to on a daily basis\citep{Pyrothechnics in automotive}
and could very well play a major role in the UAVs of tomorrow.

\subsection{Reducing impact force}

Restricted kinematic energy at impact is necessary to minimize the
risk of blunt trauma, but one must realize that it is definitely not
enough to ensure the physical integrity of people on the ground. It
doesn't take much energy to create irreversible trauma in the case
of impact of the human skull against a hard surface. As we can see
in \citep{Propeler hit video}, it doesn't take much either to perforate
human flesh with a small UAV parts. The overall geometry of the vehicle
is therefore fundamental in preventing fractures and penetrating trauma.
Ducted fans and smooth structures (i.e. without protruding parts or
with shells) could for example greatly reduce the likelihood of such
injuries. Note that a real full scale UAV collision with a human dummy
will be performed at Georgia Tech this summer to study the technical
and legal repercussions of such an incident. Following this idea,
we will discuss 3 solutions to reduce impact stress of crashing UAVs.

\subsubsection{Airbags for UAVs}

The automotive industry introduced airbags in the mid-1970s which
has had a very positive impact on the reduction of accident casualties\citep{Airbags statistics}.
It is very efficient to absorb energy during a shock and reduce the
impact force. Unsurprisingly, this technology is used to soften UAV
landing when done via a parachute like for the Elbit Systems - Skylark
II. Research has been done for this specific use of airbags\citep{UAV airbags article}
and companies are even developing dedicated products for UAV applications\citep{UAV airbags manufacturer}.
It is interesting to note that at the moment, airbags are only used
to prevent damage to the vehicle, but could be a key player in protecting
civilians as well. If asking everybody to wear a personal airbag seems
a bit out of proportion, even though it is not completely unimaginable\citep{Wearable airbag},
requiring that all UAVs carry an airbag system activated in case of
emergency would make sense. Used in conjunction with energy reducing
features, airbags may prove very efficient at minimizing human injuries,
especially because of their very fast speed of deployment\citep{UAV crach in crowd}.
Note that the deployment trigger can obviously not be the impact itself
like in cars, and that a preventive strategy is needed. Because of
the deployment speed of such systems, the minimum altitude requirement
is not as constrained as with the parachute.

\subsubsection{Engine neutralization}

As seen earlier in \citep{Propeler hit video}, propellers present
with their sharp profile and fast rotating speed are an important
danger for humans, even when the vehicle is not moving. To address
this specific threat, several passive solutions can be implemented
like ducted fans or protection shells \citep{Bouncing drone}. Furthermore,
active solution can also implemented in the same vain as controlled
disintegration, were propellers could be jettisoned in case of emergency.
This is particularly relevant in case of motor controller lockup.
Through proper design, the jettisoned propellers could use their own
shape to slow down their descent like maple keys falling from the
trees. This way, the rotational energy of the propellers is reduced
(because not attached to the rotor anymore) and they become much less
harmful to people on the ground. One obvious problem with this approach
is the fact that now you have a UAV free falling out of control. Also,
motor brakes could be implemented on the same model as for \citep{Saw stop}
but specifically designed for electric motors. Again, such technology
is relatively straight-forward and could potentially prevent severe
injuries in the future (ocular trauma for example). Finally, intelligently
cutting the propellers could be a great solution to maintain control
of the vehicle while assuring it doesn't have power to continue flying,
hence get back to the ground in a controller descent... as we will
see in \ref{subsec:The-Optimal-Active}

\subsubsection{Energy absorbing structures}

Borrowing concepts from the automotive industry, energy absorbing
structures is a key technology for minimizing trauma in case of collision.
This is achieved through proper geometrical design and material choice,
usually utilizing the various FEA analysis tools currently available.
These tools allow engineers to precisely control the way structures
will fail without relying on costly destructive testing, and for example
chose the failure points to maximize plastic deformation and energy
absorption during crash. Thanks to the recent rise of additive manufacturing
technologies, intricate polymer or metallic structures can now be
build with a single mouse click. Resistant and highly optimized structural
elements can therefore be incorporated into UAVs like porous or composite
parts (ref needed), that way providing good strength and energy absorption
at the same time. Because these features are fundamentally passive
and an integrant part of the vehicles structures, no complex electronic
mechanism is required hence making them very reliable and relatively
inexpensive compare to additional device like airbags of parachutes.

\section{Case study}

\subsection{Impact}

As UAVs get more and more capable, their weights and sizes increase
proportionally. The duality of these two factors not only represents
the likelihood of a drone impacting with a human is increasing\citep{Crash on skier},
but also the damage associated with such collision also increases.
In the state of current regulations, UAV\textquoteright s weights
can legally range between 0.5 and 55 lbs. Any object within this mass
range falling at proper velocity can cause serious injury to a human.
Furthermore, as the head is naturally exposed to vehicles falling
from the sky, blunt head trauma is one of the most likely types of
injuries. According to Knight \citep{Knight}, a human skull can be
fractured with forces as little as 73 Newtons, and with very high
probabilities at forces exceeding 510 Newtons. 

\begin{figure}[tbh]
\begin{centering}
\includegraphics[width=0.75\columnwidth]{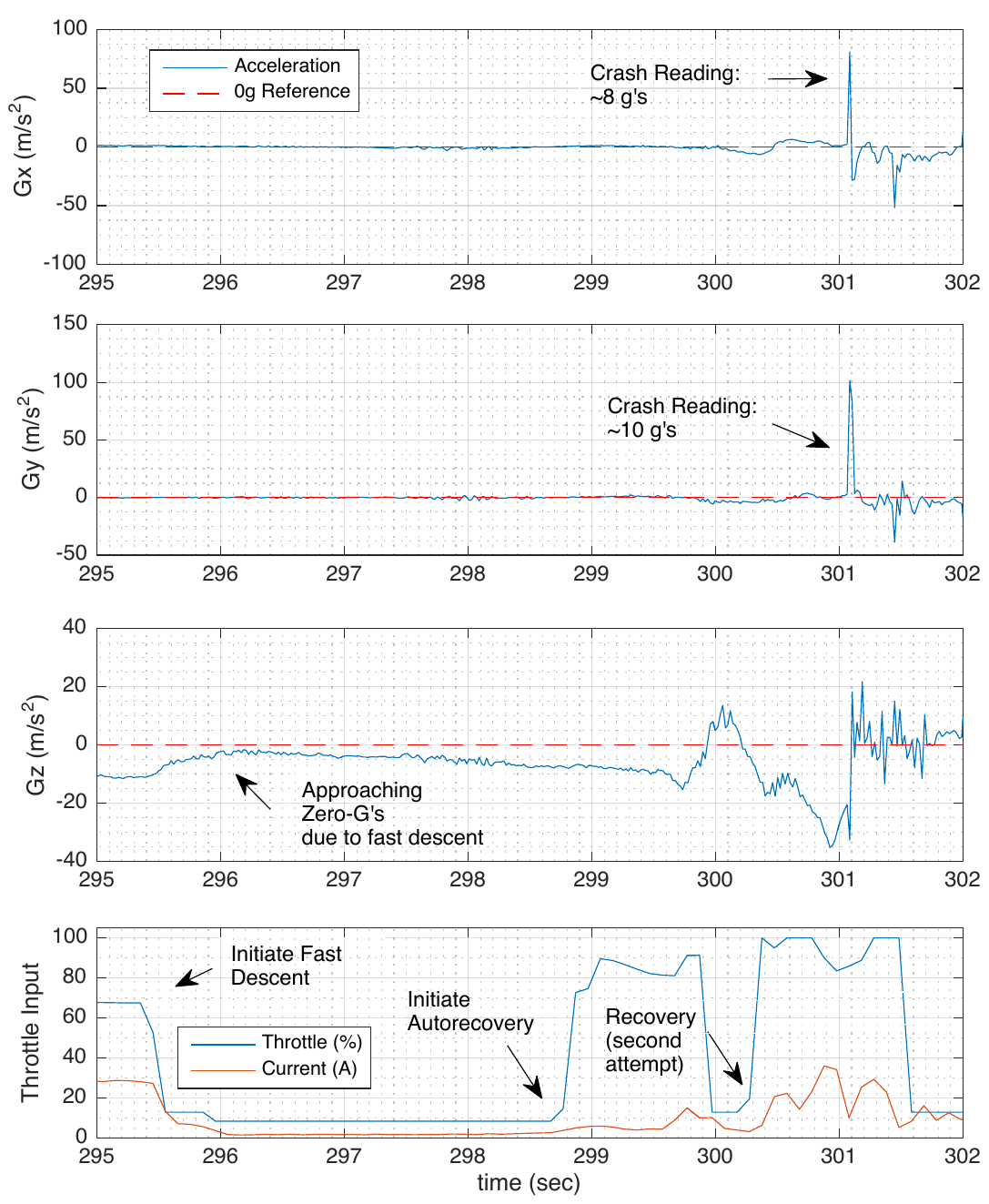}
\par\end{centering}
\caption{Filtered IMU data from the Pixhawk control board \label{fig:Filtered-IMU-data}}
\end{figure}

In the evaluation of crash data from one of Georgia Tech's fast descent-recovery
experiments, it was observed that crashes of a midsize recreational
quad-copter with a mass of 2.5 kg (5.5 lbs) can involve accelerations
exceeding 10 g\textquoteright s, implying an impact force around 250
Newtons. \ref{fig:Filtered-IMU-data} illustrates the accelerometer
data of an event in which a vehicle was not able to recover from a
fast descent, ultimately crashing into the ground below.

\begin{figure}[tbh]
\begin{centering}
\includegraphics[width=0.65\columnwidth]{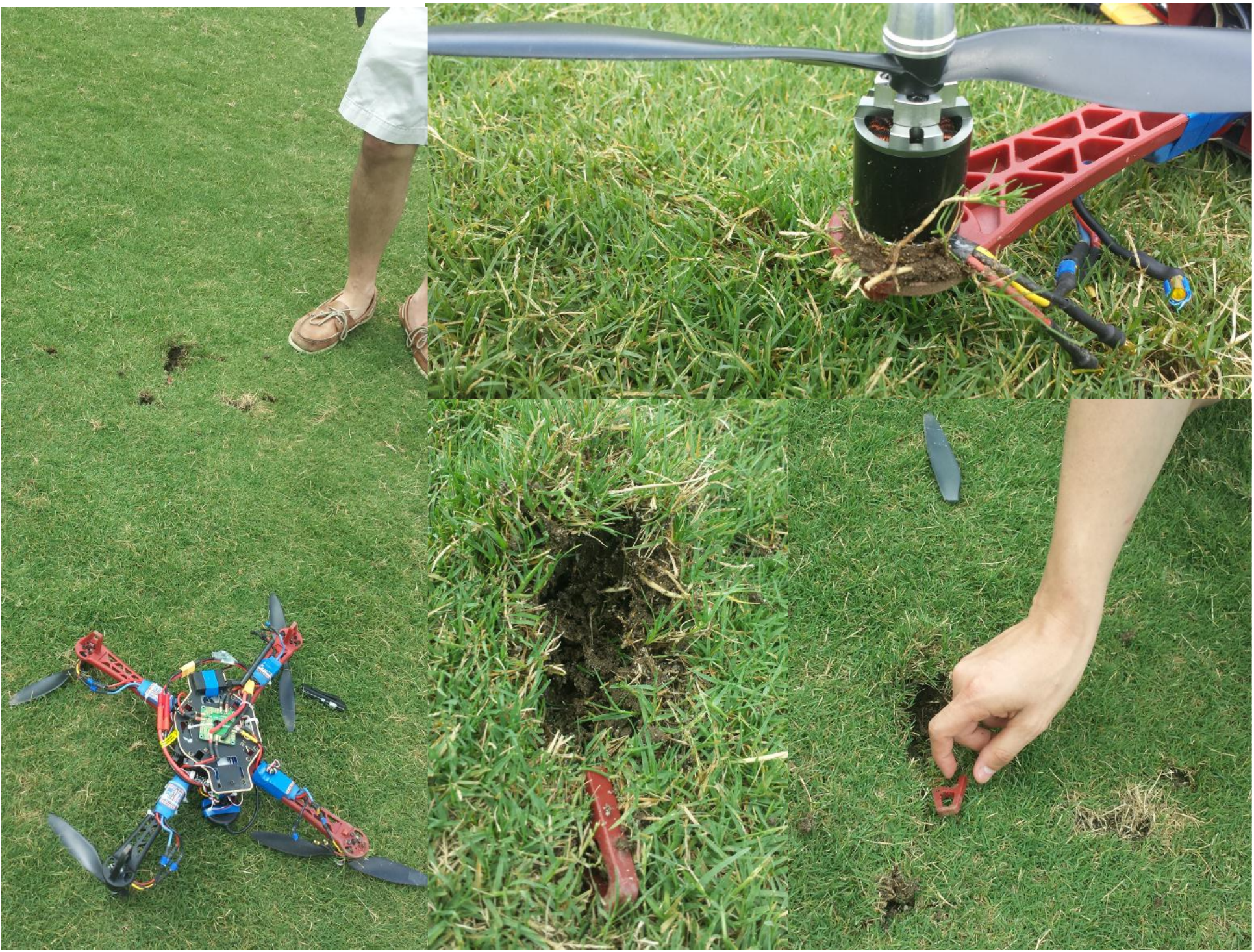}
\par\end{centering}
\caption{Forensics of a UAV crash \label{fig:Forensics-of-a} }
\end{figure}

It is noted that this is enough force to cause fractures to the skull
approximately 50\% of the time. Components such as plastic, fiberglass
spars, and landing gear pieces were shown to contain enough energy
to cause penetrating wounds when broken and can further increase that
chances of a fatal impact. Hence, it was concluded that an impact
of this type can cause a variety of life threatening injuries to the
human body. A concussion can be caused by accelerations of the head
at 8.5 m/s$^2$. Evaluations of crash data show accelerations of upwards
of 100 m/s$^2$ are very possible and a concussion from a drone strike
like this is almost certain. \ref{fig:Forensics-of-a} illustrates
the crash scene corresponding to the data shown in \ref{fig:Filtered-IMU-data}.
A video of this even can be found \href{https://www.youtube.com/watch?v=czk__30P-qA}{by clicking here.}

The ability for UAVs to cause harm is not an argument against UAVs,
but rather a reason and a framework for looking critically at UAV
safety. By understanding the potential and magnitude for bodily harm
UAVs can cause, the industry can work to reduce the likelihood of
causing injury. 

\subsection{Runaway}

As the UAV industry continues its rapid growth, it struggles to overcome
a major problem known among the enthusiastic community as \textquotedblleft fly-away.\textquotedblright{}
Fly-aways is a term used to describe a UAV that has gone wild and
flown off from its user. They are one of several safety risks that
the drone industry and aviation officials are aiming to solve. This
problem has been around for many years now, dating back to a military-drone
operation in 2010 who struggled with a 3K lb. unmanned helicopter
as it glided autonomously for 30 minutes after a software glitch severed
its connection to its U.S. Navy pilots. Furthermore, in October of
2015, U.S. Army pilots lost control of a hand-held drone over Columbus,
Ga., telling air-traffic controllers the device was headed southwest
and would run out of fuel in 40 minutes. 

Technology has made UAVs cheaper and easier to fly, giving anyone
who can spare a few hundred dollars access to small aircraft that
can climb thousands of feet. UAV\textquoteright s can zoom off or
drift away with the wind for a variety of reasons, including software
glitches, bad GPS or compass data, connection errors between receivers
and transmitters, or simple human error. Human error can occur due
to either pilot inexperience or failure to properly calibrate the
compass or configure its fail-safe functions. 

Many incidents end with the devices barreling into homes, buildings,
trees, bodies of water, and in some cases civilians \citep{UAV crach in crowd}.
Although there are not statistics on the number of fly-aways yet,
examples abound and can be found all over private YouTube channels. 

According to an article by the Wall Street Journal, a poll of 774
people who owned a popular 2.8 lb. UAV revealed that nearly a third
of the users had experienced a flyaway at some point. Furthermore,
122 of these users never saw their devices again. An extensive search
among online threads reveled that the most common cause for fly-away
drones is the \textquotedblleft return to home\textquotedblright{}
setting, a feature that returns the device to its takeoff spot in
case of a lost connection or a low battery. However, impatient users
who fly their drones before the devices have saved the \textquotedblleft home\textquotedblright{}
location often find their drones heading to a previous takeoff spot,
which could be several states away. 

\begin{figure}[tbh]
\begin{centering}
\includegraphics[width=0.75\columnwidth]{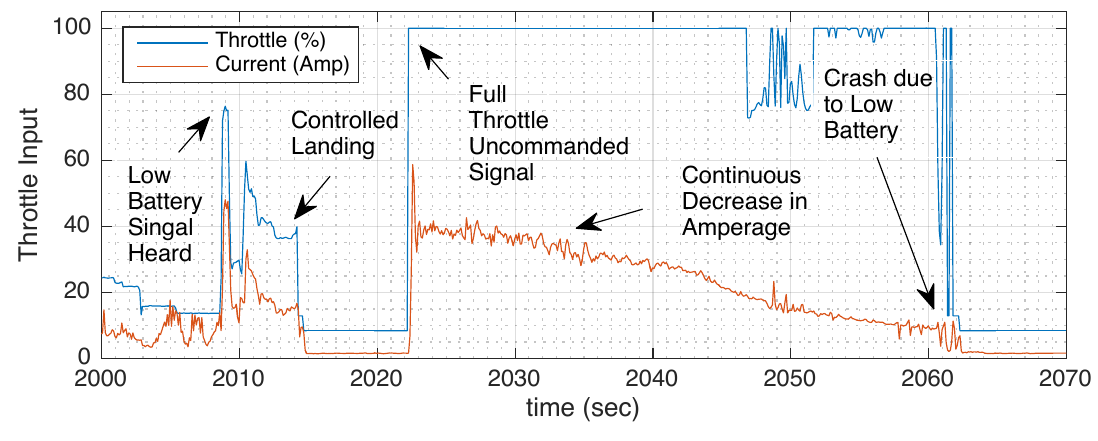}
\par\end{centering}
\caption{Control Inputs from a Run-Away Drone\label{fig:Attitude-and-Control}}
\end{figure}

Thanks to the generosity of the Afman Aeromechanic's Laboratory at
Georgia Tech, \ref{fig:Attitude-and-Control} illustrates the control
inputs from one of their own run-away drones, which occurred on one
occasion upon landing a UAV after it signaled for battery. Strangely,
the quad-rotor received an external throttle input without command
from the pilot. This caused the vehicle to rapidly accelerate upwards,
uncontrolled and unresponsive to pilot input. It can be seen that
constant throttle inputs were recorded by the Pixhawk flight controller,
despite the fact that no throttle inputs were made by the pilot. The
UAV drifted approximately 100 yards up and away from the operator
before the battery dropped below operational levels, causing it to
plummet uncontrolled to the ground in a non populated field.

Ultimately, preventing fly-away drones either requires a fix in the
core technology, or the introduction of innovative concepts. GPS and
on-board compasses, used to help orient and stabilize the devices,
can set drones adrift if tall buildings, cellphone towers or even
solar flares interfere with their accuracy. Furthermore, electromagnetic
interference, which many electrical systems emit, can also potentially
disrupt the compass and the link between a drone and its controller.
Therefore, a solution devised by this team of researchers will now
be discussed in the following chapter.

\subsection{\label{subsec:The-Optimal-Active}The Optimal Active Breaking Braking
System: an innovative safety solution for UAVs}

After having written in former paragraphs how we believe safety in
the UAV should be considered and handled, we've tried ourselves to
apply the minimum set that we deem necessary. This led to the following
requirements for a simple yet efficient solution to a safe termination
of a flight, applied here to a multi-rotor:
\begin{enumerate}
\item It should react quickly 
\item This solution shouldn't allow for an on flight or quick recovery:
when it's been activated, we can't come back to a normal operation
with a flip of a switch
\item The system should give a chance to control the speed and the path
of the descent, in order to limit the kinetic energy on impact, possibly
keep it in a given attitude (no high speed spinning, flipping) and
ideally to steer it away from people, animals or important objects
\end{enumerate}
\begin{figure}[tbh]

\centering{}\includegraphics[width=0.4\columnwidth]{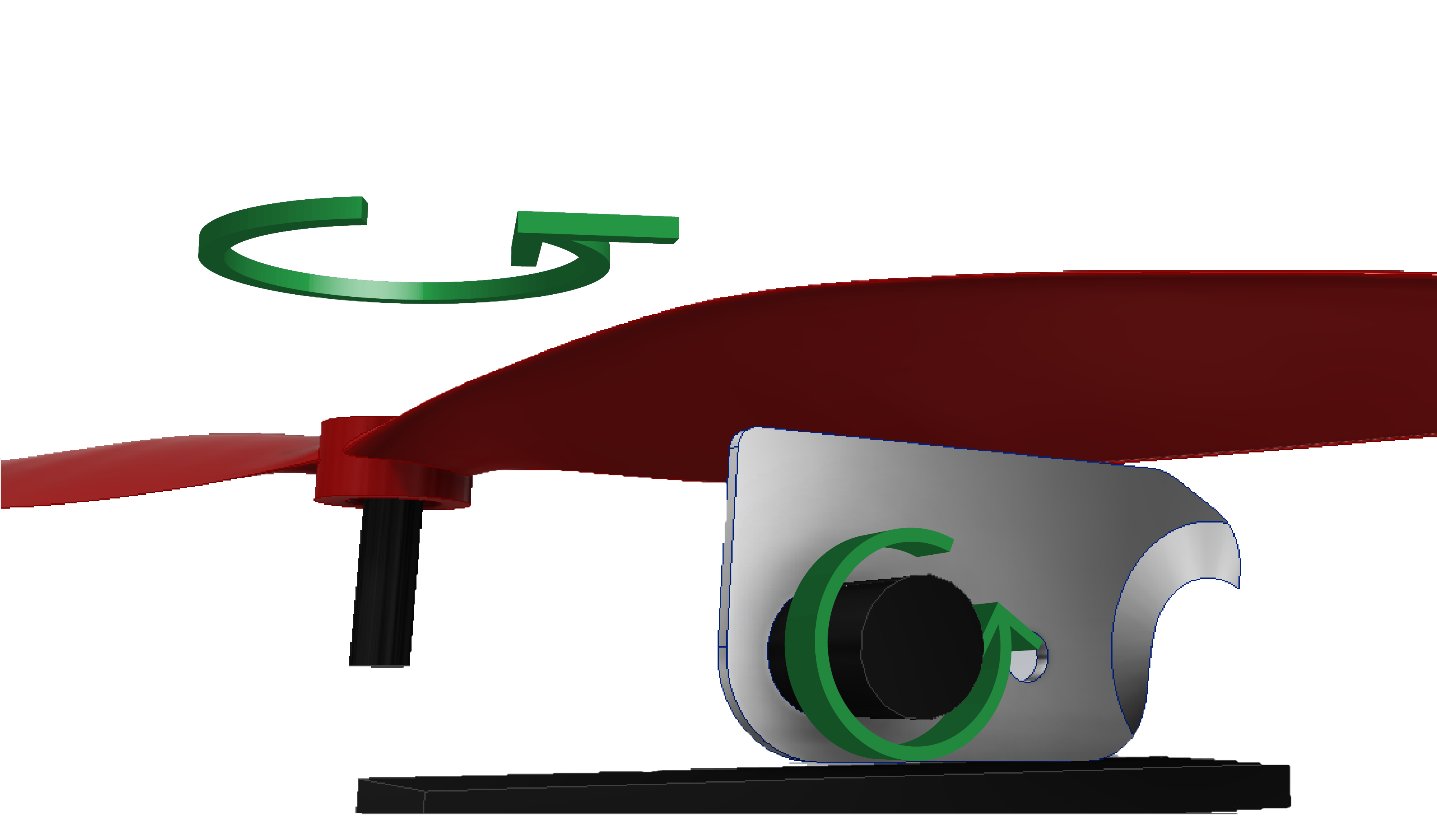}
\includegraphics[width=0.4\columnwidth]{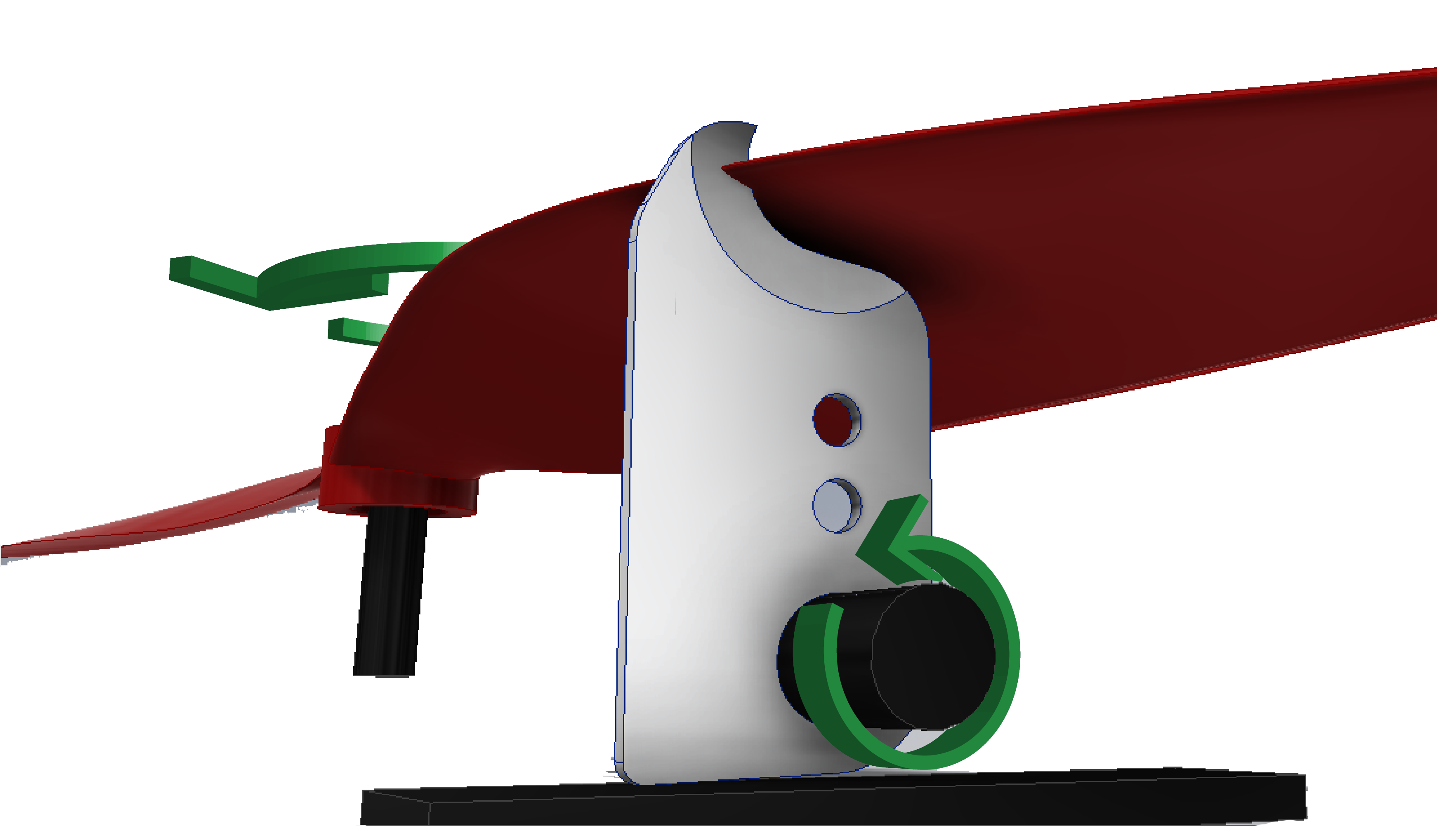}\caption{Propeller breaking tool, before and after actuation}
\end{figure}

We came up with the Optimal Active Breaking Braking System. It is
being prototyped and tested at the time of this writing, but a proof
of concept has been developed. Each arm of a quadcopter is equipped
with a servomotor that (very quickly) pushes a blade in the paths
of the rotating propellers. 

\begin{figure}[tbh]
\centering{}\includegraphics[width=0.75\columnwidth]{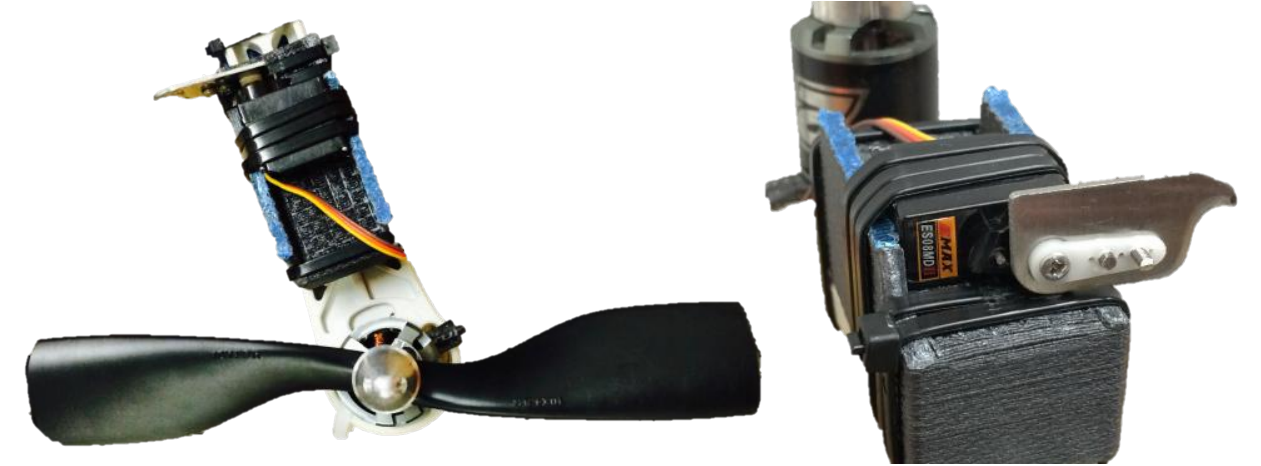}\caption{The OABBS system}
\end{figure}

Its purpose is, when activated, to cleanly cut them to an optimized
length that lowers the lift to a controlled descending speed, even
at full throttle. This leaves enough propeller surface to still be
able to guide the UAV in its landing, thus allowing to avoid people
or structures on the ground if enough control is left of the system.
Finally, the operator can't just go back to work with this UAV, he
has to change at least the propellers. A video of this system in action
is provided \href{https://www.youtube.com/watch?v=7xt9OI02JbE}{by clicking here.}

\begin{figure}[tbh]
\centering{}\includegraphics[width=0.5\columnwidth]{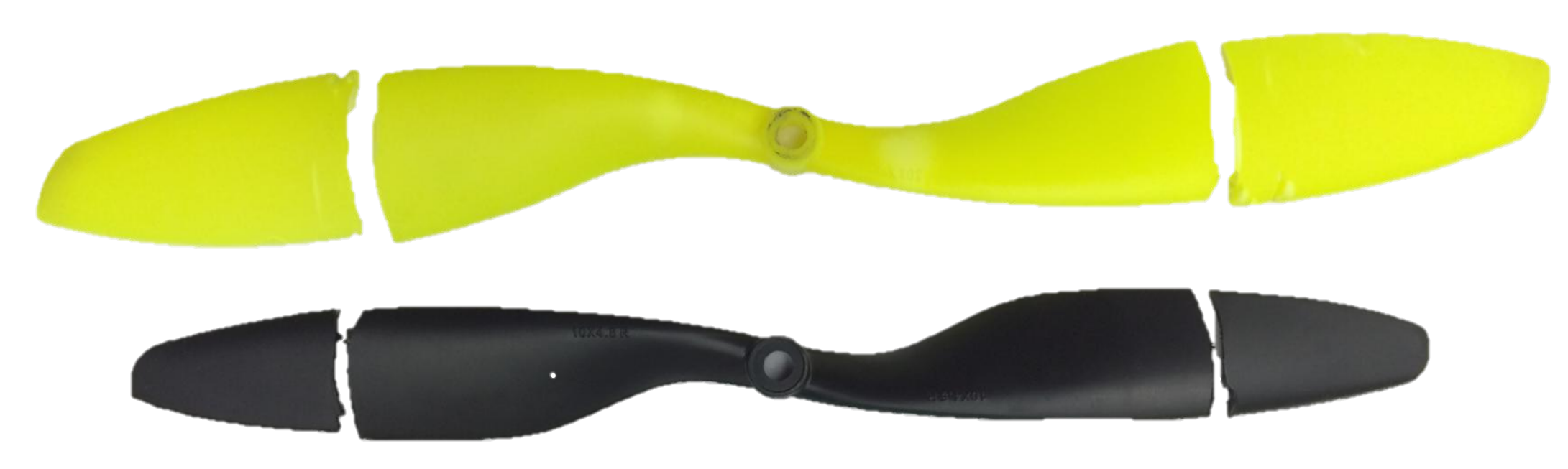}\caption{Clean cuts on different types of blade}
\end{figure}

This could also work in cooperation with a smart parachute\citep{ActiveSmartParachute}:
what's left of maneuverability can ensure it's deployed with an optimal
attitude. 

\section*{Conclusion }

This paper discusses the safety paradigm differences separating commercial
UAVs and General Aviation, placing strict emphasis in the fact that
a UAV should be crash-safe by design. Suggestions on how to manage
safety demands associated with regulatory agencies such as software
Validation and Verification as well as security related issues are
addressed from an avionics and mechanical design perspective. This
perspective includes suggestions on how to make commercial UAVs a
reality while maintaining software validation and verification costs
in mind, which have the potential to cripple the entire UAV revolution.
Furthermore, a mechanical design perspective explores several solutions
on how one can reduce the danger associated with failed devices from
a design perspective, with special emphasis on reducing impact energy.
Lastly, a case study, using real crash data from the Afman Aeromechanics
Laboratory at Georgia Tech, evaluates the energy of an impact and
suggests a novel mechanical solution that is introduced in this paper
for the first time. This novel device aims at tackling the single
biggest issue encountered by UAV pilots to date, safely landing a
run away UAV.

\end{document}